\documentclass[conference]{IEEEtran}
\IEEEoverridecommandlockouts
\usepackage{cite}
\usepackage{longtable}
\usepackage{amsmath,amssymb,amsfonts}
\usepackage{algorithmic}
\usepackage{graphicx}
\usepackage{textcomp}
\usepackage{xcolor}

\def\BibTeX{{\rm B\kern-.05em{\sc i\kern-.025em b}\kern-.08em
    T\kern-.1667em\lower.7ex\hbox{E}\kern-.125emX}}
\begin{document}

\title{Towards Automated Machine Learning: Evaluation and Comparison of AutoML 
Approaches and Tools
}

\author{
\IEEEauthorblockN{Anh Truong\IEEEauthorrefmark{1}, 
Austin Walters\IEEEauthorrefmark{1}, 
Jeremy Goodsitt\IEEEauthorrefmark{1}, 
Keegan Hines\IEEEauthorrefmark{1}, 
C. Bayan Bruss\IEEEauthorrefmark{1}, 
Reza Farivar\IEEEauthorrefmark{1}\IEEEauthorrefmark{2}}
\and\and
\IEEEauthorblockA{
	\hspace{1.2cm}\IEEEauthorrefmark{1}\textit{Applied Research, Center for Machine Learning} \\
	\hspace{1.2cm}\textit{Capital One}\\
	\hspace{1.2cm}McLean, VA, USA \\
	\hspace{1.2cm}\{anh.truong, austin.walters, jeremy.goodsitt \\
	\hspace{1.2cm}keegan.hines, bayan.bruss\}@capitalone.com}
\and
\IEEEauthorblockA{
	\IEEEauthorrefmark{2}\textit{Department of Computer Science} \\
	\textit{University of Illinois}\\
	Urbana-Champaign, IL, USA \\
	farivar2@illinois.edu
	}
}

\maketitle

\begin{abstract}
There has been considerable growth and interest in industrial applications of machine learning (ML) in recent years. ML engineers, as a consequence, are in high demand across the industry, yet improving the efficiency of ML engineers remains a fundamental challenge. Automated machine learning (AutoML) has emerged as a way to save time and effort on repetitive tasks in ML pipelines, such as data pre-processing, feature engineering, model selection, hyperparameter optimization, and prediction result analysis. In this paper, we investigate the current state of AutoML tools aiming to automate these tasks. We conduct various evaluations of the tools on many datasets, in different data segments, to examine their performance, and compare their advantages and disadvantages on different test cases.
\end{abstract}


\section{Introduction}
Automated Machine Learning (AutoML) promises major productivity boosts for data scientists, ML engineers and ML researchers by reducing repetitive tasks in machine learning pipelines. 
There are currently a number of different tools and platforms (both open-source and commercially available solutions) that try to automate these tasks. The goal of this paper is to address the following questions: (i) what are the available ML functionalities provided by the tools; (ii) how the tools perform when facing a wide spectrum of real world datasets; (iii) find the trade-off between optimization speed and accuracy of the results; and (iv) the reproducibility of the results (a.k.a. tool robustness). 

 The rest of the paper is organized as follows. Section \ref{sec:background} covers the history and background of AutoML tools. Next, in Section \ref{sec:pipeline} we compare the tools' features and functionalities on an automated ML pipeline including data preprocessing, model selection, hyperparameter optimization, and model interpretation. After that, in Section \ref{sec:experiment} we experimentally evaluate the performance of a selected subset of these tools on a large variety of datasets and a range of supervised ML tasks. Finally, we conclude the paper in Section \ref{sec:conclusion}.

\section{Background and History}\label{sec:background}
Between 1995 to 2015 many ML libraries and tools were developed, spanning from Weka (1990s), RapidMiner (2001), Scikit-learn (2007-2010), H2O (2011), and Spark MLlib (2013) among many others. Deep Neural Network platforms have also gained popularity in the last 5 years. Tensorflow (2015), Keras (2015) and MXNet (2015) contributed to the wide adoption of deep learning models. 

During this time period it became evident to many ML practitioners that extracting the best performance from machine learning models requires substantial human expertise. Developing good models from a dataset is almost an art form involving intuition, experience, and many tedious manual tasks to tune algorithmic parameters. 
The combination of market pressure for more ML engineers, and the tedious nature of developing `optimal' ML solutions sparked the idea of automating the ML tasks. 

AutoML's initial effort came out of academia and ML practitioners first, and later startups. One of the first attempts was Auto-Weka (2013)\cite{Kotthoff:autoweka} from Universities of British Columbia (UBC) and Freiburg, which utilizes algorithms provided by Weka \cite{weka}. Auto-sklearn (2014) \cite{autosklearn} came next from the University of Freiburg. TPOT \cite{tpot} was developed at the University of Pennsylvania (2015). Auto-ml \cite{auto-ml}, an open-source python package, was released in 2016 (to avoid confusion with the general term `AutoML', please note the spelling for this tool). Auto-sklearn, Auto-ml, and TPOT are all built on the well-known `scikit-learn' ML package. Other tools followed, including Auto-Keras (2017) \cite{autokeras} from the Texas A$\&$M University running on top of Keras, Tensorflow and Scikit-learn. MLjar (2018) \cite{mljar} also uses Scikit-learn, in conjunction with Tensorflow. On the same timeframe, some startups started developing their tools for AutoML. Datarobot \cite{datarobot}, \cite{datarobot1}, \cite{datarobot2} launched its automated machine learning tool in 2015. H2O-Automl \cite{h2o}, \cite{h2o1} was introduced by the H2O (2016), using ML models from the H2O platform. The H2O team later released their commercialized H2O-DriverlessAI product (2017) \cite{h2odriverlessai}, and SparkCognition introduced Darwin (2018) \cite{darwin} utilizing their own ML platform.

After a while, the large cloud providers and technology companies followed suit, offering Automated Machine Learning as a Service (AMLaaS) or standalone products. Google Cloud Automl (2017) \cite{gcp} runs on Google Cloud platform. Microsoft AzureML (2018) \cite{azureml} takes advantage of algorithms on Azure, and Salesforce's TransmogrifAI (2018) \cite{transmogriai} runs on top of Spark ML, and Uber's Ludwig (2019) \cite{ludwig} runs model training on Horovod, Uber's open-source distributed training framework.

The aforementioned platforms emphasize different aspects of the AutoML space. For example, Darwin, H2O-DriverlessAI and DataRobot provide the functionality of detecting and processing time-series data. They also offer interactive UI to help customers experiment quickly with different machine learning tasks. H2O-DriverlessAI exports a Plain Old Java Object (POJO) or a Model ObJect Optimized (MOJO) for the optimized models to be easily deployed in any Java-supported platform. TPOT exports optimized code for developers. Auto-ml offers `categorical ensembling', where segments of categories in a column can have different models. Google Cloud AutoML and Auto-keras conduct neural network search \cite{Zoph:NAS}, \cite{pham:NAS}, for both image and text data.

\section{AutoML platforms' features and functionality comparison: The common pipeline}\label{sec:pipeline}
 \begin{figure}[!htb]
\begin{center}
\centerline{\includegraphics[width=\columnwidth]{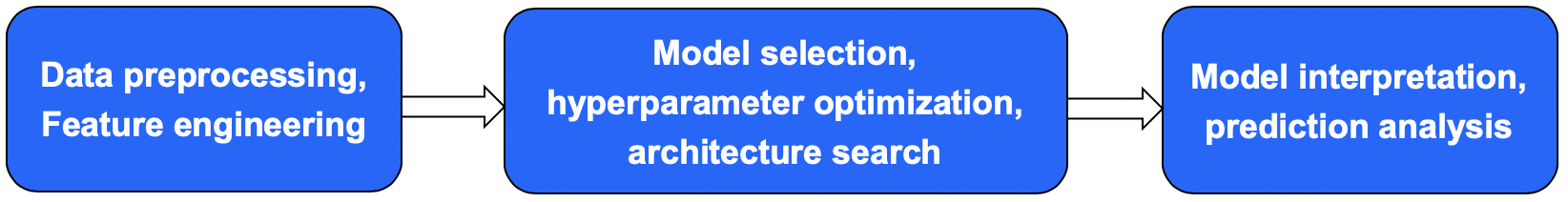}}
\caption{The common AutoML pipeline.}
\label{fig:pipeline}
\end{center}
\end{figure}
\begin{figure*}[!ht]
\begin{center}
\centerline{\includegraphics[width=0.99\textwidth]{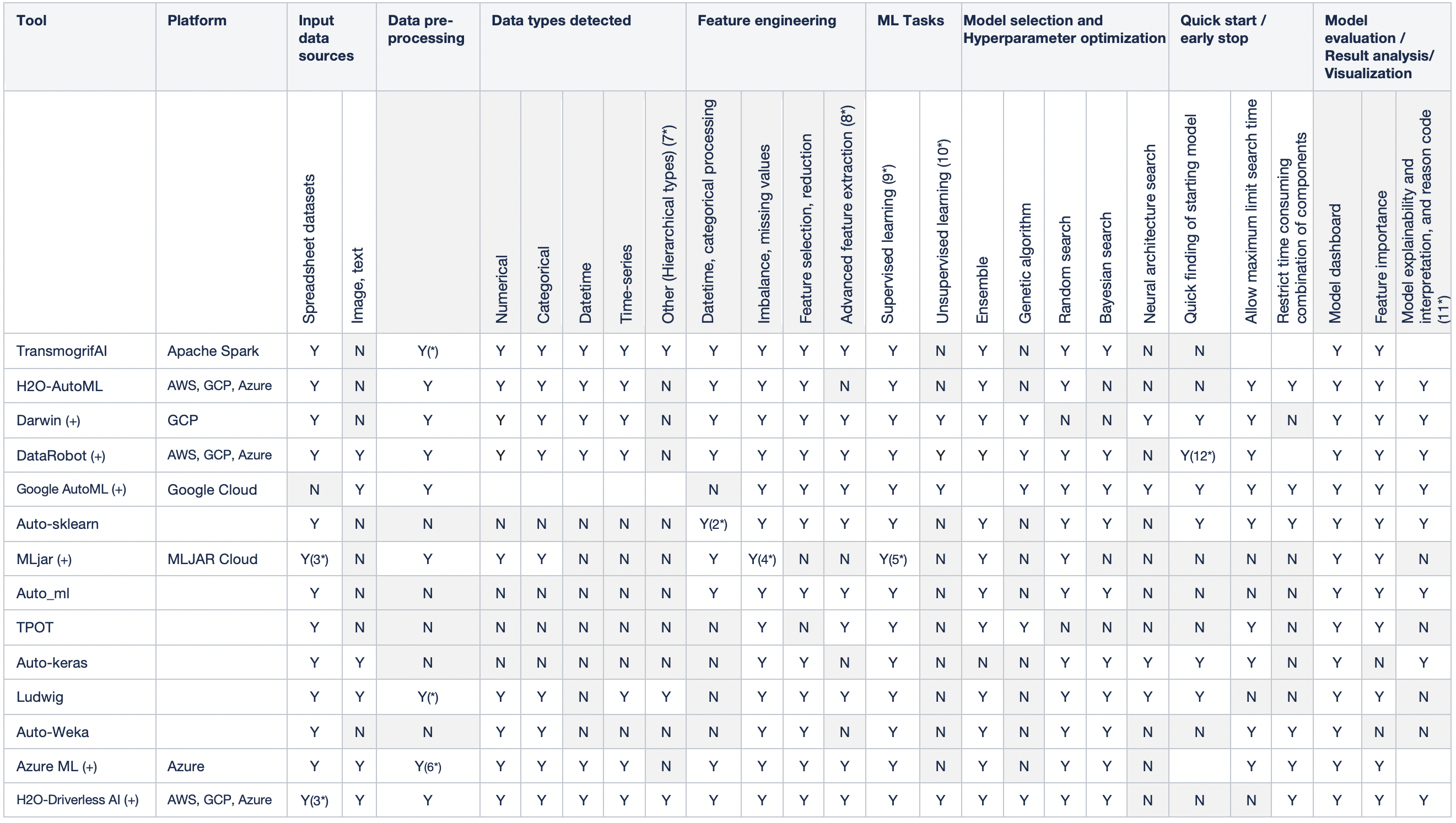}}
\caption{\footnotesize Comparison table of functionality for AutoML tools. $(+)$: commercialized tools; $(^{*})$: the function is not very stable, it fails for some datasets; $(2^{*})$: categorical input must be converted into integers; $(3^{*})$: datasets have to include headers; $(4^{*})$: missing values must be represented as NA; $(5^{*})$: multiclass classification not provided; $(6^{*})$: need some users' input for dataset description such as column types; $(7^{*})$: ability to detect primitive data types and rich data types such as: text (id, url, phone), numerical (integer, real); $(8^{*})$: advanced feature processing: bucketing of values, removing features with zero variance or features with drift over time; $(9^{*})$: supervised learning includes binary classification, multiclass classification, regression; $(10^{*})$: unsupervised learning includes clustering and anomaly detection; $(11^{*})$: model interpretation and explainability refers to techniques such as LIME, Shapley, Decision Tree Surrogate, Partial Dependence, Individual Conditional Expectation, Lift chart, feature fit, prediction distribution plot, accuracy over time, hot spot and reason codes; $(12^*)$: confirmed by a company spokesperson, we could not find public documentation at the time of publication; In a few empty cells, it is not clear that the functionality is provided from documentations of the tools, to the best of our knowledge.
}
\label{fig:table-functionality}
\end{center}
\end{figure*}

Most AutoML tools follow a common three stage pipeline illustrated in Figure \ref{fig:pipeline}. In general, these three components are optimized iteratively to obtain the best outcome. Figure \ref{fig:table-functionality} briefly summarizes the comparison across the surveyed tools. More detailed comparisons follow in the subsequent sections.
 
\subsection{Data Preprocessing and Feature Engineering}
Data preprocessing is typically the first task in ML pipelines. At the moment, this task is not handled very well by any of the AutoML tools and still requires considerable human intervention. In particular, this task requires data type and schema detection which have not been widely supported among the AutoML tools. However, once data types are identified, the tools provide appropriate feature engineering for the next component in the pipeline. TransmogrifAI seems to be further ahead in this regard by supporting different detailed data types detection (e.g., addresses, phone numbers, names, currency, etc), however this functionality appears to not be very stable on multiple datasets. H2O-Automl, H2O-DriverlessAI, DataRobot, MLjar and Darwin gain some advantage by offering the ability to detect basic data types or schemas, currently limited to numerical, categorical and time-series data. Auto-ml, Auto-sklearn, AzureML and Ludwig are less favorable here, in the sense that they can only do feature engineering from user-input specifications, e.g. data types for each column. The other tools need much more human interaction on feature engineering. Auto-sklearn requires users input to convert categorical data into integers (e.g., using label encoder) before any other transformation. 
TPOT and Auto-keras do not provide either data preprocessing or feature generation steps and instead require users to manually perform data pre-processing, and only accept numerical feature matrices.

\subsection{Model Selection, Hyperparameter Optimization, and Architecture Search}\label{sec:model}
In this step, the extracted features from the previous step are used to train many different types of models, each with many different sets of parameters (hyperparameter optimization), then the best model (or an ensemble of models) is selected as the final model. 
Each tool supports a collection of existing machine learning algorithms to build models. They include, but not limited to, Logistic Regression, tree-based algorithms, SVM, and neural network models. H2O-Automl, Ludwig, DataRobot, Darwin, Auto-ml, Auto-sklearn, MLjar, TransmogrifAI, and TPOT all work in this fashion for supervised methods. DataRobot, H2O-DriverlessAI and Darwin provide additional unsupervised methods such as clustering and outlier detection. TPOT and Darwin also utilize genetic algorithms to iteratively select the best traits of each model and pass them to the next generation. Google Cloud AutoML and Auto-keras work differently, utilizing neural architecture search to select the best neural network model. 

For hyperparameter optimization, some of the most popular methods are grid search, random search, and Bayesian search. Auto-Weka uses SMAC (Sequential Model-based Algorithm Configuration, \cite{smac}) while Auto-sklearn utilizes SMAC3, a re-implementation of SMAC to efficiently perform Bayesian optimization. H2O-Automl and MLjar apply random search on the parameter spaces, while H2O-DriverlessAI, Ludwig, Auto-ml, TransmogrifAI and Auto-keras use both random and Bayesian search. 

In order to reduce time for model search and hyperparameter optimization, it is common to prune the  parameter space. In the first approach, the tools attempt to quickly find an initial parameter set. Auto-sklearn and Darwin use pre-processed `meta-features' from previously trained datasets, each with a known `meta-learner'. Given a target dataset, they find a similar dataset based on `meta-feature', and use the closest 'meta-learners' as the initial model. 
The second approach is to use the relationship between model selection and hyperparameter optimization. H2O-Automl uses the combination of random grid search with stacked ensembles, as diversified models improve the accuracy of ensemble method. 
The third approach is to fix an allowed runtime for the tools to search for a best model. All AutoML tools, except Auto-ml, currently offer this option.
The fourth approach (only applies for H2O-Automl and Auto-sklearn) is to restrict the parameters that cause a slow optimization. For example, non-linear feature approximation combined with KNN models is restricted as it dramatically slows down the optimization.

\subsection{Model Interpretation and Prediction Analysis}
This component is currently applied to most commercialized tools such as H2O-DriverlessAI, DataRobot and Darwin whereas it is not the concentration of non-commercialized tools. In essence, it provides detailed result representation through model dashboards, feature importance and different visualization methods, e.g., lift chart and prediction distribution. Those tools even highlight outlier data points that the best model was not confident in predictions, and support `reason code', LIME, Shapley, and partial dependence, etc., for better model interpretation.

\section{Experimental evaluation}\label{sec:experiment}
We evaluate a selected subset \footnote{Due to the unavailability of the licensed or trial versions, we have not evaluated most commercialized tools. Some other open-sourced tools have not been evaluated due to the lack of widely supported Python wrappers.} of AutoML tools on nearly 300 datasets collected from Openml \cite{openml}, which allows users to query data for different use cases. Detailed descriptions on the datasets are given on the Table \ref{tab:datasets} in the Appendix. 
\begin{figure}[!htb]
\begin{center}
\centerline{\includegraphics[width=0.85\columnwidth]{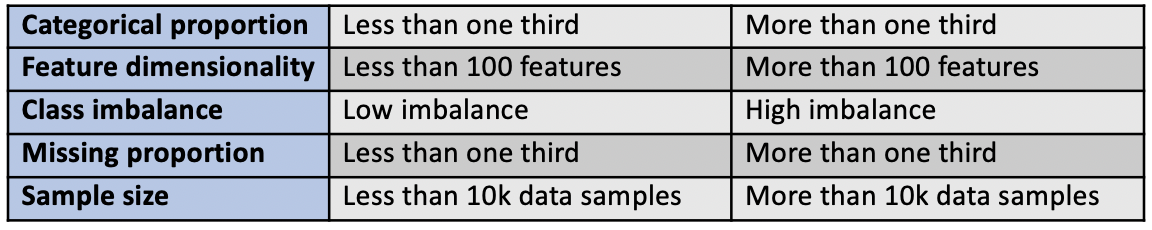}}
\caption{\footnotesize Data segments used for evaluation. Each cell is referred to as a `data segment'. For example, in the first row, \textit{'Less than one third'} stands for datasets with the categorical proportion less than $1/3$.}
\label{fig:table-data}
\end{center}
\end{figure}
The two advantages of using Openml datasets are: (i) the datasets are already pre-processed into the numerical features \footnote{Although there are still a few datasets containing text or non-numerical features, those are not included in this paper.}, therefore the same data will be fed to all AutoML tools, minimizing the risk of bias from data selection process; and (ii) guarantee a fair comparison among the tools as some do not provide the pre-processing steps for raw datasets. In order to evaluate AutoML tools on a variety of dataset characteristics, we selected multiple datasets according to the criteria depicted in Figure \ref{fig:table-data}. 
\begin{figure*}[!htb]
\begin{center}
\centerline{\includegraphics[width=0.99\textwidth]{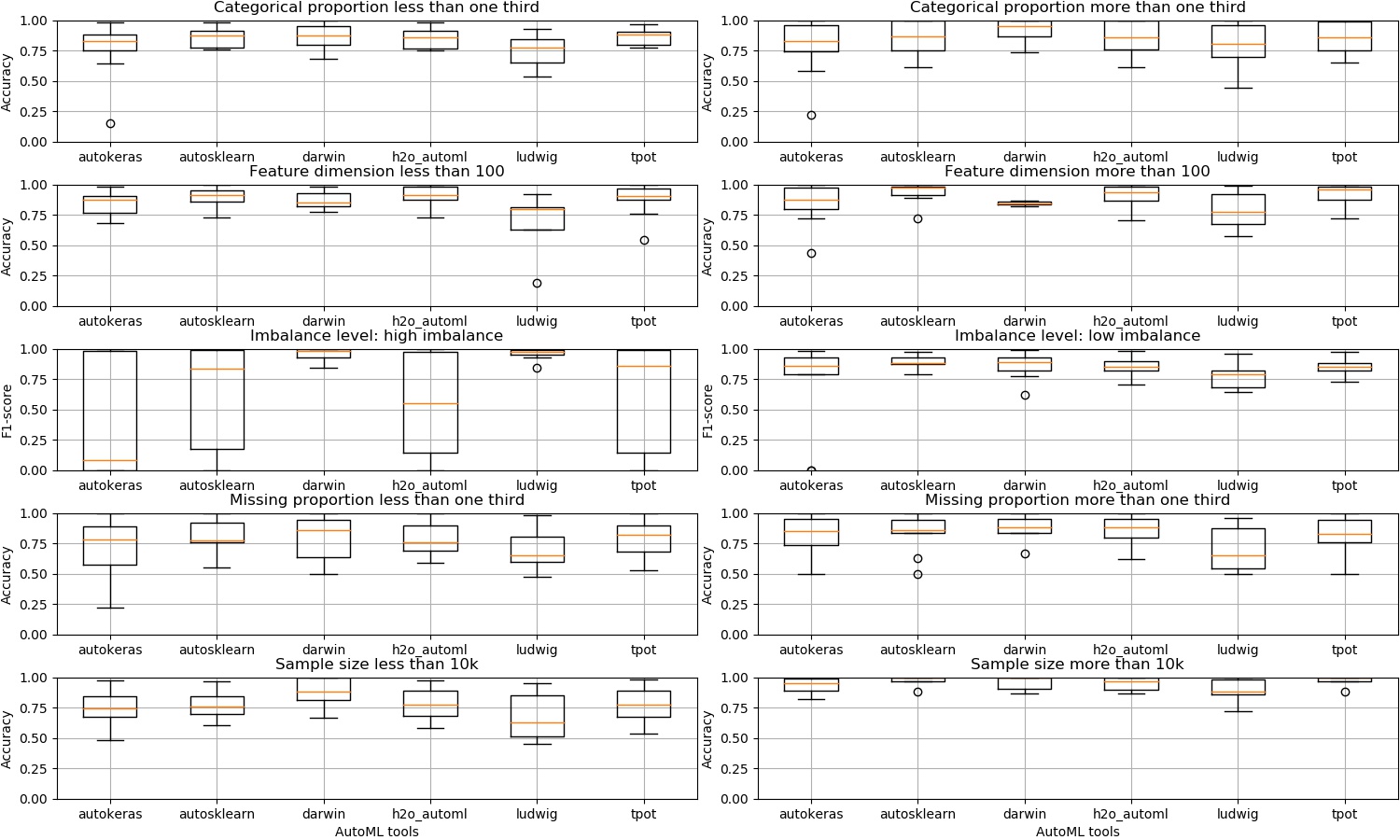}}
\caption{\footnotesize Evaluation of AutoML tools on binary classification task across ten data segments (depicted in Figure \ref{fig:table-data}). Each diagram refers to a data segment. All experiments are run up to 15 minutes. Some experiments are completed faster but in some other cases, several tools cannot obtain results after that time limit. Specifically, the percentage of experiments that did not finish in 15 minutes are: Ludwig $4\%$, H2O-Automl $8\%$, TPOT $13\%$,  Darwin $26\%$, Auto-sklearn $30\%$}
\label{fig:manydata-bin}
\end{center}
\end{figure*}
For the sake of clarity, each cell in this table is referred to as a \underline{`data segment'}, each containing datasets with different sample sizes, feature dimensions, categorical features ratio (defined as the ratio of number of categorical features over total number of features), missing proportion (proportion of samples with at least one missing feature), and class imbalance (samples in minority class vs. in majority class). Each dataset is divided into two parts, one for training and another for testing with the ratio $4:1$. All AutoML tools are applied to the same training and testing proportions of all datasets. For all evaluations, the following tools and associated versions are used: Darwin 1.6, Auto-sklearn 0.5.2, Auto-keras 0.4.0, Auto-ml 2.9.10, Ludwig 0.1.2, H2O-Automl 3.24.0.5, TPOT 0.10.1.

In the next subsections, we will evaluate AutoML tools on different test cases, each with three different supervised learning tasks: binary classification, multiclass classification, and regression. All experiments are run on Amazon EC2 p2.xlarge instances, which provide 1 Tesla K80 GPU, 4 vCPUs (Intel Xeon E5-2686, 2.30Ghz) and 61 GiB of host memory. 

Setting a time-limit for all experiments is not straightforward. On the one hand, we would like to let each tool run as long as it takes to produce the best results. On the other hand, with 3 ML tasks, 300 datasets and 6 tools, we have 5,400 experiments to run. To keep the experiment run-time and cost to practical limits, we aim for a `completion target' of 70$\%$, i.e., we select a run-time for which all tools are able to finish the AutoML tasks for 70$\%$ of the datasets. All the AutoML tools proved to be capable of hitting the 70$\%$ target within 15 minutes for binary classification. 5 out of the 6 tools (all but Darwin) hit 70$\%$ target for regression, and 4 out of 6 tools hit the 70$\%$ target in multiclass classification, (TPOT nearly reaches the target, and Darwin misses the target again). As Darwin appears to be slow in convergence, and to be fair to the other tools, it is excluded from our completion target analysis. We therefore decide to run all our extensive experiments (5,400) for 15 minutes time-limits, for a total of 1,350-hour EC2 run-time (which includes the overhead of benchmark harness code), where the results are detailed in Section \ref{sec:manydata}. We then run another experiment with a randomized subset of our datasets for longer time limits to evaluate the performance of the tools when more time is given to finish. The results of this latter experiment (3 tasks, 5 data segments, 6 run-time periods, 7 tools, for a total of 717 EC2 hours including benchmark harness overhead) are detailed in Section \ref{sec:time-limit}.
Note that the Auto-ml tool was not included in the extensive experiments as it does not offer an option to limit its run-time from a user-input value (15 minutes in our case), it simply can only run to completion. As such, its results are only included among the experiments in Section \ref{sec:time-limit}.
\begin{figure*}[!htb]
\begin{center}
\centerline{\includegraphics[width=0.99\textwidth]{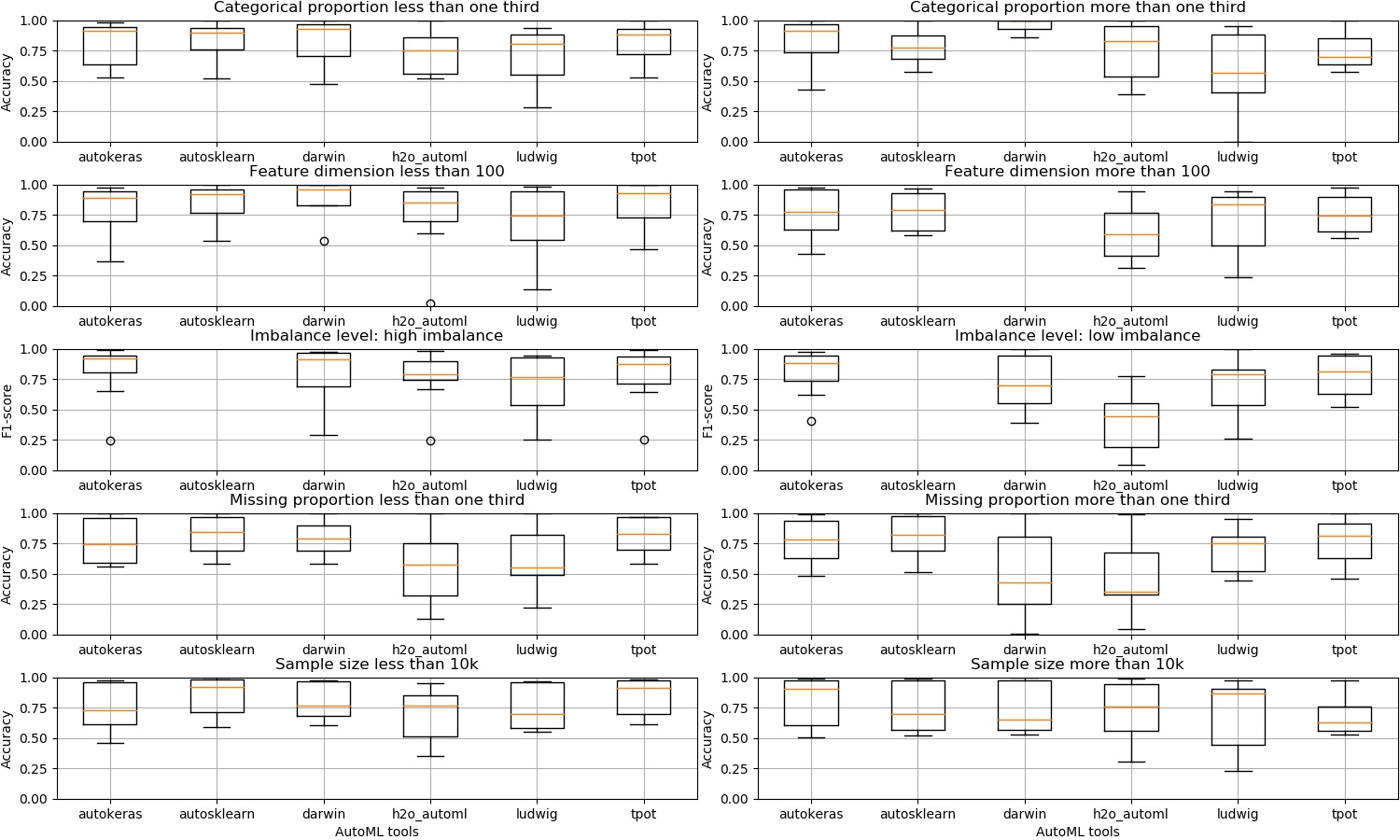}}
\caption{\footnotesize Evaluation of AutoML tools on multiclass classification task across ten data segments (depicted in Figure \ref{fig:table-data}). Each diagram refers to a data segment. All experiments are run up to 15 minutes. Some experiments are completed faster but in some other cases, several tools cannot obtain results after that time limit. Specifically, the percentage of experiments that did not finish in 15 minutes are: Auto-keras $2\%$, H2O-Automl $21\%$, Ludwig $24\%$, Auto-sklearn $30\%$, TPOT $35\%$,  Darwin $51\%$.}
\label{fig:manydata-mul}
\end{center}
\end{figure*}
\subsection{Evaluation on multiple data segments}\label{sec:manydata}
In this section, we investigate the performance of the tools across many datasets and applications (please see Table \ref{tab:datasets} in the Appendix for the detailed descriptions on the datasets). To that end, the evaluated data is divided into ten segments (as shown in Figure \ref{fig:table-data}), each including ten random datasets. `Accuracy' is the comparison metric used for binary and multiclass classification tasks and `Mean Squared Error (MSE)' is used for regression tasks.

Figure \ref{fig:manydata-bin} shows the performance of AutoML tools for binary classification task in different data segments. In this Figure, the performance is represented in the box-whisker format, where each box shows the median, and the first and third quartiles of the  performance at the two ends. Note that, for the data segment with class imbalance (third row in Figure \ref{fig:manydata-bin}), F1-score is used instead of the regular accuracy as it is a more appropriate metric for imbalanced data.

It can be observed from Figure \ref{fig:manydata-bin} that, the performance of AutoML tools fluctuate more with a larger number of categorical features, and fluctuate less with more data samples. This makes intuitive sense, as the tools will learn better with more data samples, and each tool has different approaches to encode categorical values that result in different performance. In addition, most tools suffer from the imbalanced datasets except Ludwig and Darwin. Comparing the tools against each other, H2O-Automl and Darwin slightly outperform the rest, however it is worth reiterating that   Darwin cannot deliver results for $26\%$ of all datasets. Auto-sklearn and TPOT perform slightly worse than the aforementioned tools. Auto-keras does not perform as well as other tools for most datasets in binary classification. As noted before, in this experiment, we limit the optimization time to 15 minutes.
Here, TPOT manages to complete and deliver results within the 15-minutes time limit for $87\%$ of datasets, while Darwin and Auto-sklearn suffer slightly higher non-delivering ratios of $26\%$ and $30\%$ respectively. Ludwig's performance appears to fluctuate the most compared to other tools.

The performance of the tools for muticlass classification is illustrated in Figure \ref{fig:manydata-mul}.
Here, minimal variation was found when evaluating between data segments of the same categories (the two graphs in each row). For this multiclass classification task, Auto-keras and Auto-sklearn slightly outperform the rest, even though Auto-sklearn cannot deliver results within the time limit for $30\%$ of datasets. TPOT comes next after these two tools. Finally, Ludwig, H2O-Automl and Darwin perform slightly worse than the rest.

\begin{figure*}[!htb]
\begin{center}
\centerline{\includegraphics[width=0.99\textwidth]{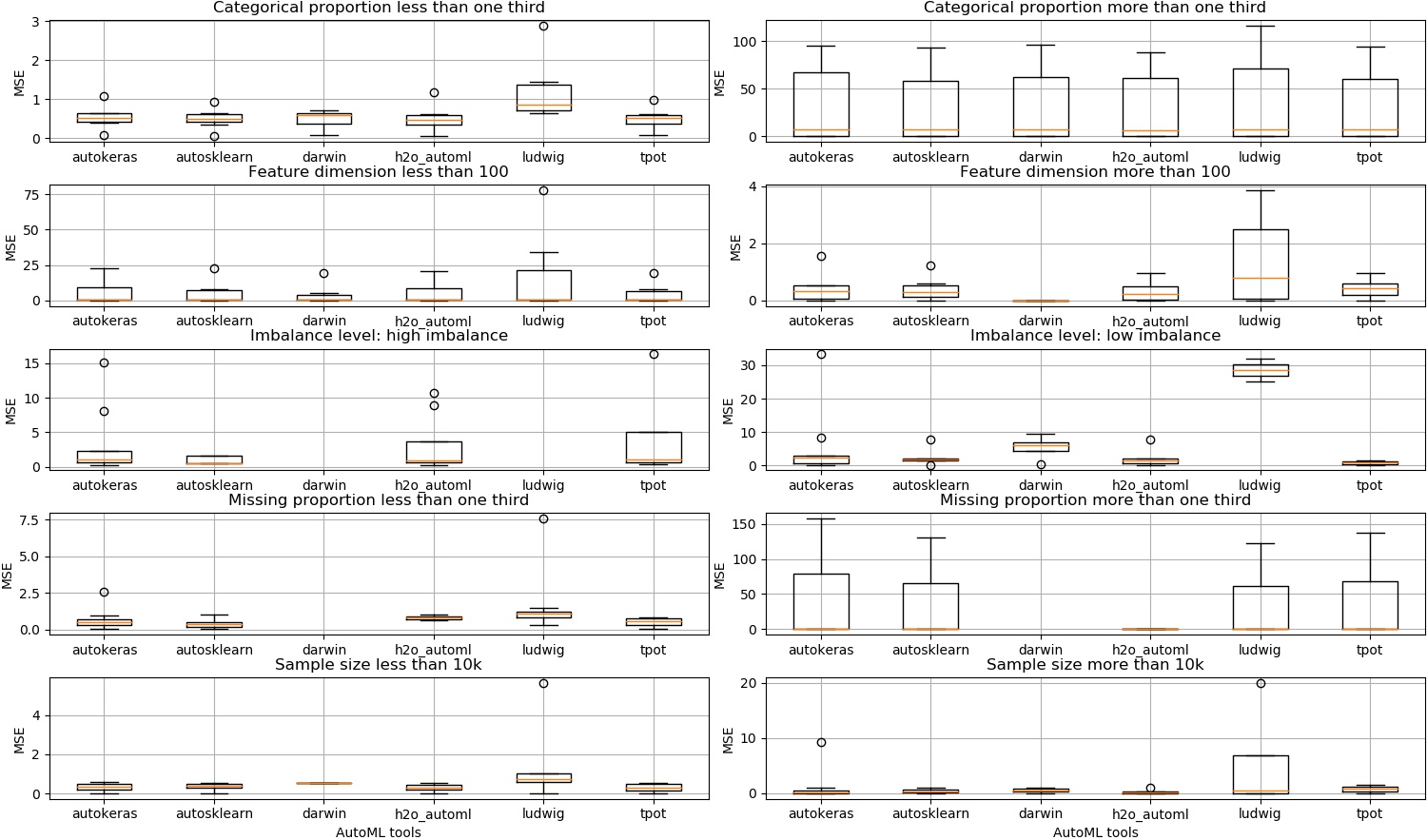}}
\caption{\footnotesize Evaluation of AutoML tools on regression task across ten data segments (depicted in Figure \ref{fig:table-data}). Each diagram refers to a data segment. All experiments are run up to 15 minutes. Some experiments are completed faster but in some other cases, several tools cannot obtain results after that time limit. Specifically, the percentage of experiments that did not finish in 15 minutes are: Auto-keras $4\%$, H2O-Automl $11\%$, Auto-sklearn $20\%$, Ludwig $24\%$, TPOT $25\%$,  Darwin $56\%$}
\label{fig:manydata-reg}
\end{center}
\end{figure*}

Figure \ref{fig:manydata-reg} shows the performance of the tools for regression task. 
The results from this task has similar trends to binary classification on categorical features. Furthermore, the performance variance tends to increase for all tools when the features dimensions decreases, or missing proportion increases. For this task, H2O-Automl and Auto-sklearn slight outperform Auto-keras and TPOT while Darwin cannot deliver results on half of the datasets.

To summarize what we have seen from the three different ML tasks, Auto-keras does not perform as well as other tools for some datasets in binary classification. In other words, whether Auto-keras can perform well or not (in binary classification) depends significantly on the nature of the dataset. For multiclass classification, H2O-Automl performs slightly worse than the rest. For the regression task, Auto-keras, H2O-Automl and Auto-sklearn outperform the rest for most data segments (even though Auto-sklearn struggles somewhat more to complete results in the allotted 15 minutes, failing in $26\%$ of datasets). TPOT performs slightly worse than those three tools, Ludwig's performance varies across the datasets, and Darwin can only complete work on about half of the datasets in the allotted 15 minutes.

\subsection{Evaluation on time limit}\label{sec:time-limit}
\begin{figure*}[!htb]
\begin{center}
\centerline{\includegraphics[width=0.99\textwidth]{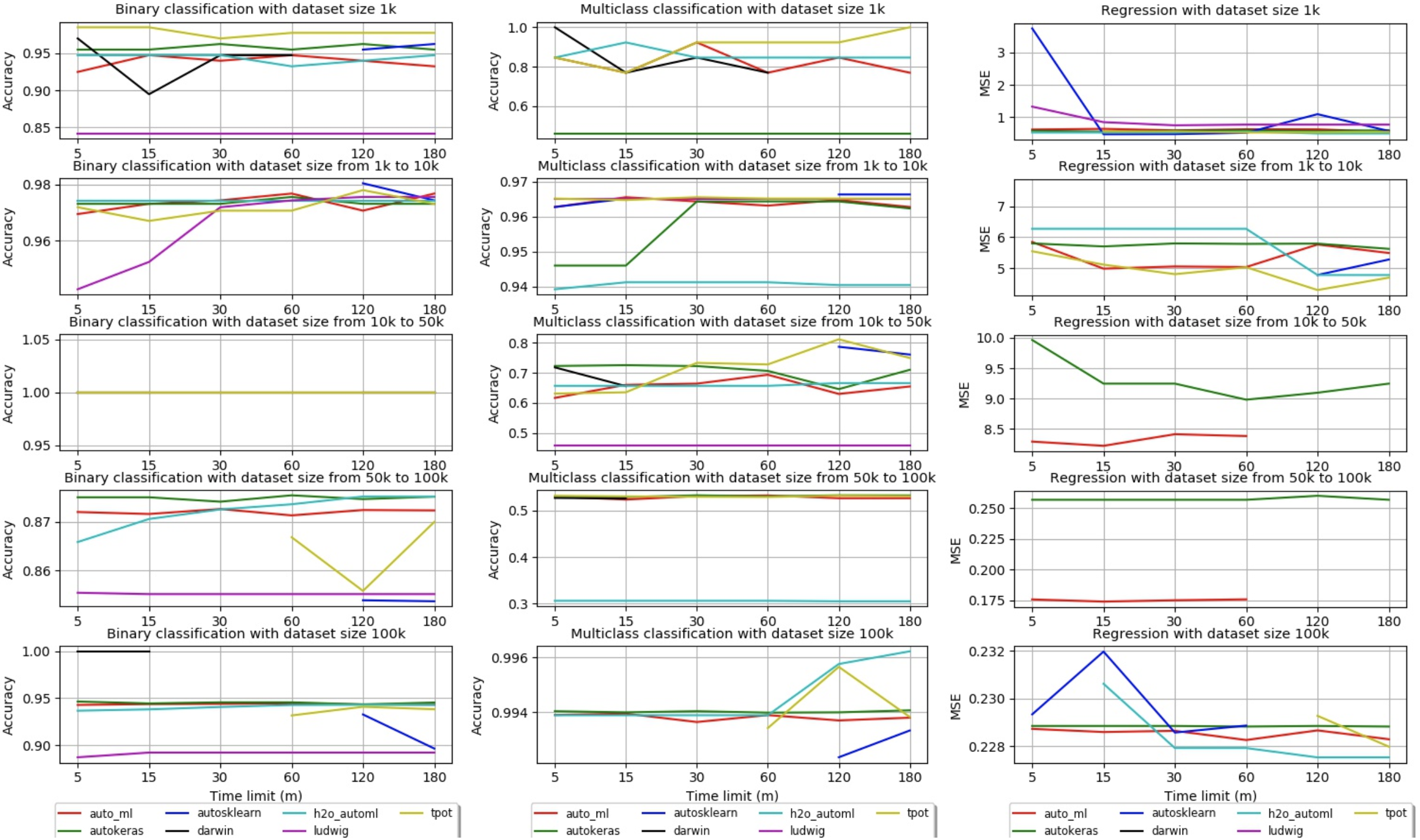}}
\caption{\footnotesize Evaluation of AutoML on multiple time limits. The left (middle) subgraphs show the accuracy of tools for binary (multiclass) classification. The right subgraphs show the mean squared error of tools for regression. From top-to-bottom: each row shows a random dataset in the increasing order of the sample size, from 1000 to 100000. Note: In the left graph of the third row, all tools except Darwin obtain the same performance although the graph displays only the result for TPOT; in the graphs at the rows 3 $\&$ 4, column 3, all tools except Auto-keras and Auto-ml cannot deliver results due to the large number of features, roughly $62000$ and $21000$, respectively; in the second graph of third column, we omit the results of Ludwig as its error is roughly $100$-times larger than the others.}
\label{fig:timelimit}
\end{center}
\end{figure*}

Our next targeted evaluation is to explore the impact of time limit in order to investigate how quickly the tools can deliver the results, and whether the tools can consistently guarantee better results given more time availability. We performed various time-limit experiments for datasets with different sample sizes. Here, we randomly select a dataset given a sample size range (i.e., we pick a uniformly random dataset among all datasets in each sample size range) and evaluate each tools' accuracy bounded by the time limits: 5 minutes, 15 minutes, 30 minutes, 1 hour, 2 hours and 3 hours. 
Since the dataset sizes do not exceed one million samples, the maximum allotted time of 3 hours should allow the tools to converge. Figure \ref{fig:timelimit} shows the results of this evaluation. As observed from the figure, most tools can generally improve the performance (increase the accuracy for classification tasks and decrease the mean-squared error for regression task) given more time for their optimization. Among the tools, H2O-Automl, Auto-keras and Ludwig converge to the optimal performance very quickly for most cases, roughly within 15 minutes. Auto-sklearn needs almost 2-3 hours to obtain reasonable results while TPOT converges slightly faster. Darwin appears fluctuating its performance even with more time for optimization.

\begin{figure*}[!htb]
\begin{center}
\centerline{\includegraphics[width=0.99\textwidth]{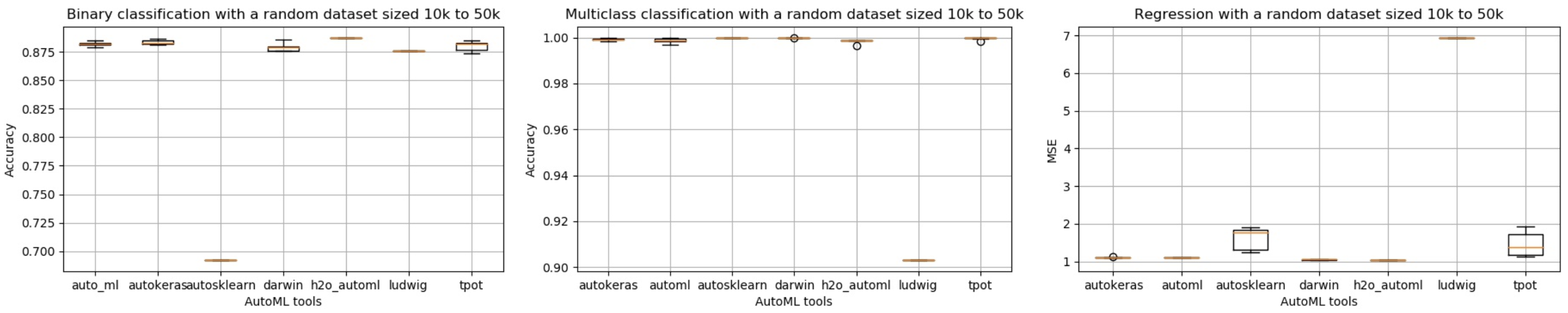}}
\caption{Evaluation of AutoML tools on robustness.}
\label{fig:robustness}
\end{center}
\end{figure*}

\subsection{Evaluation on robustness}
In this evaluation, we test the robustness of AutoML tools, i.e., whether the tools deliver similar results across multiple runs on the same input datasets. For each task, we select a random dataset with the sample size from 10000 to 50000 (this is a common sample size for many real-world datasets) and run each tool on it for ten different times, each times in 10 minutes. The results are illustrated in Figure \ref{fig:robustness}. We observe that H2O-Automl and Ludwig obtain very stable performance across three different tasks. Darwin, Auto-keras and Auto-ml get slightly less stable performance than H2O-Automl. TPOT and Auto-sklearn are somewhat unstable in regression task. It is worth noting that even though Ludwig's performance is very stable, it deviates largely from others.

\section{Conclusions and Future Work}\label{sec:conclusion}
In this paper, we have evaluated AutoML tools on their capabilities in the common machine learning pipeline. At the current state, different tools have different approaches for model selection and hyperparameter optimization. Commercialized tools such as H2O-DriverlessAI, DataRobot and Darwin extend their offering functionality on the first and the third component of the pipeline where they are able to detect the data schema, run feature engineering, and analyze the detailed results for interpretation purpose. In contrast, open source tools focus more on the second task in the pipeline, which is training and selecting the best model itself. 

In addition, we have evaluated tools across many datasets on different data segments. We observed that most AutoML tools obtain reasonable results in terms of their performance across many datasets. However, there is no perfect tool at the current state yet, no tool managed to outperform all others on a plurality of tasks. Across the various evaluations and benchmarks we have tested, H2O-Automl, Auto-keras and Auto-sklearn performed better than Ludwig, Darwin, TPOT and Auto-ml. In particular, H2O-Automl slightly outperforms the rest for binary classification and regression, and quickly converges to the optimal results. However, it suffers from low performance in multiclass classification. Auto-keras is very stable across all tasks and performs slightly better than the rest for multiclass classification and ties with H2O-Automl for regression, but suffers from low performance in binary classification. For a production environment where the computation speed and performance stability are key requirements, these two tools might be good candidates depending on applications and machine learning tasks. Auto-sklearn ties with H2O-Automl and Auto-keras for all tasks but it is comparatively slower than the other two and usually requires longer run time. Other tools such as Ludwig, Darwin, TPOT and Auto-ml showed more varying results depending on the dataset and task. 

Ultimately there is no one AutoML tool at this point that can clearly outperform every other tool. We are at an early juncture for Automated Machine Learning, and there are many innovations announced at a rapid pace. We believe as the tools mature and borrow ideas from each other, they will gain more strength in their core task. We also observed a gap in the AutoML tools' support for the first and third stages of the AutoML pipeline, and expect major developments to happen in those areas in near future.
\\
\\
	\textit{\textbf{Disclaimer}: For commercialized tools, our analyses and descriptions are consistent with our understanding derived from publicly available documentation and product descriptions. In some of these cases, we are unable to explore source code and regret any factual errors that may arise.}

\textit{A subsidiary of Capital One - Capital One Ventures - is an investor in H2O.ai. During the course of our research we were neither in contact with H2O.ai nor Capital One Ventures.}
\nocite{*}
\bibliographystyle{IEEEtran}
\bibliography{IEEEabrv,ref}

\appendix
\begin{table*}[!htbp]
\tiny
\centering
\caption{Dataset descriptions.}
\begin{tabular}{*6c}
\hline
\multicolumn{2}{c}{Binary classification}  &  \multicolumn{2}{c}{Multiclass classfication} & \multicolumn{2}{c}{Regression}\\
\hline
Id   & Name   & Id    & Name   &   Id   &   Name\\
954	&	spectrometer	&	342	&	 squash-unstored	&	3584	&	 QSAR-TID-12665\\
23499	&	 breast-cancer-dropped-missing-attributes-values	&	385	&	 tr31.wc	&	3536	&	 QSAR-TID-12868\\
862	&	 sleuth-ex2016	&	48	&	 tae	&	3682	&	 QSAR-TID-100790\\
905	&	 chscase-adopt	&	1565	&	 heart-h	&	4096	&	 QSAR-TID-30028\\
724	&	 analcatdata-vineyard	&	1516	&	 robot-failures-lp1	&	4057	&	 QSAR-TID-10547\\
40978	&	 Internet-Advertisements	&	1535	&	 volcanoes-b5	&	197	&	 cpu-act\\
983	&	 cmc	&	40708	&	 allrep	&	3394	&	 QSAR-TID-20154\\
40649	&	 GAMETES-Heterogeneity-20atts-1600-Het-0.4-0.2-50-EDM-2-001	&	40476	&	 thyroid-allhypo	&	573	&	 cpu-act\\
41156	&	 ada	&	181	&	 yeast	&	1028	&	 SWD\\
40648	&	 GAMETES-Epistasis-3-Way-20atts-0.2H-EDM-1-1	&	28	&	 optdigits	&	558	&	 bank32nh\\
959	&	 nursery	&	1044	&	 eye-movements	&	1594	&	 news20\\
881	&	 mv	&	41163	&	 dilbert	&	1583	&	 w3a\\
977	&	 letter	&	40926	&	 CIFAR-10-small	&	564	&	 fried\\
734	&	 ailerons	&	1536	&	 volcanoes-b6	&	344	&	 mv\\
357	&	 vehicle-sensIT	&	119	&	" BNG(cmc, nominal, 55296)"	&	1578	&	 real-sim\\
1111	&	 KDDCup09-appetency	&	255	&	 BNG(cmc)	&	1588	&	 w8a\\
1112	&	 KDDCup09-churn	&	554	&	 mnist-784	&	1591	&	 connect-4\\
41150	&	 MiniBooNE	&	23397	&	 COMET-MC-SAMPLE	&	273	&	 IMDB.drama\\
1212	&	 BNG(SPECTF)	&	148	&	" BNG(zoo, nominal, 1000000)"	&	1191	&	 BNG(pbc)\\
1369	&	" BNG(kr-vs-kp, 1000, 1)"	&	157	&	 RandomRBF-10-1E-3	&	1192	&	 BNG(autoHorse)\\
875	&	 analcatdata-chlamydia	&	455	&	 cars	&	533	&	 arsenic-female-bladder\\
472	&	 lupus	&	1413	&	 MyIris	&	556	&	 analcatdata-apnea2\\
461	&	 analcatdata-creditscore	&	1528	&	 volcanoes-a2	&	689	&	 chscase-vine2\\
796	&	 cpu	&	11	&	 balance-scale	&	551	&	 analcatdata-michiganacc\\
37	&	 diabetes	&	1499	&	 seeds	&	678	&	 visualizing-environmental\\
1590	&	 adult	&	41169	&	 helena	&	4553	&	 TurkiyeStudentEvaluation\\
1059	&	 ar1	&	4535	&	 Census-Income	&	287	&	 wine-quality\\
1442	&	 MegaWatt1	&	188	&	 eucalyptus	&	1191	&	 BNG(pbc)\\
917	&	 fri-c1-1000-25	&	453	&	 analcatdata-bondrate	&	608	&	 fri-c3-1000-10\\
806	&	 fri-c3-1000-50	&	1361	&	" BNG(anneal.ORIG, 1000, 5)"	&	1572	&	 german.numer\\
742	&	 fri-c4-500-100	&	40670	&	 dna	&	1586	&	 w6a\\
1116	&	 musk	&	41082	&	 USPS	&	315	&	 us-crime\\
1563	&	 dbworld-subjects-stemmed	&	313	&	 spectrometer	&	422	&	 topo-2-1\\
40910	&	 Speech	&	1548	&	 autoUniv-au4-2500	&	1584	&	 w4a\\
357	&	 vehicle-sensIT	&	41166	&	 volkert	&	412	&	 Phen\\
1157	&	 AP-Endometrium-Kidney	&	1082	&	 rsctc2010-6	&	3626	&	 QSAR-TID-103437\\
1123	&	 AP-Endometrium-Breast	&	1078	&	 rsctc2010-2	&	3363	&	 QSAR-TID-19905\\
4134	&	 Bioresponse	&	41163	&	 dilbert	&	3574	&	 QSAR-TID-10958\\
1164	&	 AP-Endometrium-Uterus	&	391	&	 re0.wc	&	3586	&	 QSAR-TID-104390\\
41159	&	 guillermo	&	1514	&	 micro-mass	&	3955	&	 QSAR-TID-12406\\
734	&	 ailerons	&	1086	&	 ovarianTumour	&	3266	&	 QSAR-TID-11169\\
41142	&	 christine	&	40996	&	 Fashion-MNIST	&	3836	&	 QSAR-TID-142\\
1448	&	 KnuggetChase3	&	1544	&	 volcanoes-e3	&	40753	&	 delays-zurich-transport\\
1021	&	 page-blocks	&	477	&	 fl2000	&	3707	&	 QSAR-TID-25\\
868	&	 fri-c4-100-25	&	377	&	 synthetic-control	&	4050	&	 QSAR-TID-218\\
990	&	 eucalyptus	&	1539	&	 volcanoes-d2	&	620	&	 fri-c1-1000-25\\
749	&	 fri-c3-500-5	&	1087	&	 hepatitisC	&	3230	&	 QSAR-TID-10930\\
162	&	 SEA(50000)	&	1523	&	 vertebra-column	&	223	&	 stock\\
874	&	 rabe-131	&	1117	&	 desharnais	&	3298	&	 QSAR-TID-12163\\
1130	&	 OVA-Lung	&	277	&	 meta-ensembles.arff	&	1093	&	 Brainsize\\
950	&	 arsenic-female-lung	&	41000	&	 jungle-chess-2pcs-endgame-panther-elephant	&	200	&	 pbc\\
804	&	 hutsof99-logis	&	155	&	 pokerhand	&	534	&	 cps-85-wages\\
43	&	 haberman	&	380	&	 SyskillWebert-Bands	&	665	&	 sleuth-case2002\\
963	&	 heart-h	&	457	&	 prnn-cushings	&	515	&	 baseball-team\\
40536	&	 SpeedDating	&	379	&	 SyskillWebert-Goats	&	553	&	 kidney\\
992	&	 analcatdata-broadway	&	148	&	" BNG(zoo, nominal, 1000000)"	&	31	&	 credit-g\\
960	&	 postoperative-patient-data	&	184	&	 kropt	&	203	&	 lowbwt\\
72	&	 BNG(kr-vs-kp)	&	4541	&	 Diabetes130US	&	213	&	 pharynx\\
1010	&	 dermatology	&	1363	&	" BNG(anneal.ORIG, 5000, 1)"	&	224	&	 breastTumor\\
40680	&	 mofn-3-7-10	&	130	&	 BNG(segment)	&	301	&	 ozone-level\\
40647	&	 GAMETES-Epistasis-2-Way-20atts-0.4H-EDM-1-1	&	400	&	 tr41.wc	&	3915	&	 QSAR-TID-11109\\
804	&	 hutsof99-logis	&	1520	&	 robot-failures-lp5	&	3668	&	 QSAR-TID-100071\\
1462	&	 banknote-authentication	&	1113	&	 KDDCup99	&	3925	&	 QSAR-TID-101090\\
714	&	 fruitfly	&	1404	&	" BNG(lymph, 1000, 10)"	&	3760	&	 QSAR-TID-117\\
719	&	 veteran	&	1356	&	" BNG(anneal, 5000, 10)"	&	193	&	 bolts\\
726	&	 fri-c2-100-5	&	186	&	 braziltourism	&	1432	&	 colon-cancer\\
933	&	 fri-c4-250-25	&	74	&	" BNG(letter, nominal, 1000000)"	&	3833	&	 QSAR-TID-12898\\
808	&	 fri-c0-100-10	&	1177	&	 BNG(primary-tumor)	&	619	&	 fri-c4-250-50\\
904	&	 fri-c0-1000-50	&	375	&	 JapaneseVowels	&	3625	&	 QSAR-TID-11000\\
40994	&	 climate-model-simulation-crashes	&	401	&	 oh10.wc	&	3522	&	 QSAR-TID-100629\\
55	&	 hepatitis	&	171	&	 primary-tumor	&	200	&	 pbc\\
56	&	 vote	&	327	&	 bridges	&	222	&	 echoMonths\\
172	&	 shuttle-landing-control	&	328	&	 bridges	&	232	&	 fishcatch\\
757	&	 meta	&	40966	&	 MiceProtein	&	516	&	 pbcseq\\
802	&	 pbcseq	&	2	&	 anneal	&	524	&	 pbc\\
1114	&	 KDDCup09-upselling	&	5	&	 arrhythmia	&	566	&	 meta\\
1000	&	 hypothyroid	&	7	&	 audiology	&	231	&	 hungarian\\
470	&	 profb	&	57	&	 hypothyroid	&	315	&	 us-crime\\
38	&	 sick	&	378	&	 ipums-la-99-small	&	1072	&	 qqdefects-numeric\\
51	&	 heart-h	&	381	&	 ipums-la-98-small	&	41021	&	 Moneyball\\
40713	&	 dis	&	71	&	" BNG(anneal.ORIG, nominal, 1000000)"	&	383	&	 tr45.wc\\
1142	&	 OVA-Endometrium	&	1086	&	 ovarianTumour	&	384	&	 tr21.wc\\
951	&	 arsenic-male-lung	&	388	&	 tr23.wc	&	385	&	 tr31.wc\\
1018	&	 ipums-la-99-small	&	39	&	 ecoli	&	387	&	 tr11.wc\\
40910	&	 Speech	&	279	&	 meta-stream-intervals.arff	&	388	&	 tr23.wc\\
1056	&	 mc1	&	183	&	 abalone	&	395	&	 re1.wc\\
977	&	 letter	&	40498	&	 wine-quality-white	&	397	&	 tr12.wc\\
1001	&	 sponge	&	1402	&	" BNG(lymph, 1000, 1)"	&	398	&	 wap.wc\\
450	&	 analcatdata-lawsuit	&	30	&	 page-blocks	&	400	&	 tr41.wc\\
1045	&	 kc1-top5	&	1541	&	 volcanoes-d4	&	40596	&	 slashdot\\
803	&	 delta-ailerons	&	1554	&	 autoUniv-au7-500	&	31	&	 credit-g\\
920	&	 fri-c2-500-50	&	61	&	 iris	&	273	&	 IMDB.drama\\
739	&	 sleep	&	1548	&	 autoUniv-au4-2500	&	300	&	 isolet\\
1486	&	 nomao	&	40996	&	 Fashion-MNIST	&	386	&	 oh15.wc\\
1167	&	 pc1-req	&	46	&	 splice	&	390	&	 new3s.wc\\
1125	&	 AP-Omentum-Prostate	&	6	&	 letter	&	392	&	 oh0.wc\\
787	&	 witmer-census-1980	&	1493	&	 one-hundred-plants-texture	&	393	&	 la2s.wc\\
481	&	 biomed	&	1385	&	" BNG(letter, 10000, 5)"	&	394	&	 oh5.wc\\
935	&	 fri-c1-250-10	&	1041	&	 gina-prior2	&	396	&	 la1s.wc\\
763	&	 fri-c0-250-10	&	147	&	" BNG(waveform-5000, nominal, 1000000)"	&	399	&	 ohscal.wc\\
\hline
\end{tabular}
\label{tab:datasets}
\end{table*}

\end{document}